\documentclass[accepted]{uai2021} 

\usepackage[american]{babel}

\usepackage{natbib} 
\usepackage{mathtools} 
\usepackage{booktabs} 
\usepackage{tikz, pgfplots} 
\usepackage{graphicx}
\usepackage{subcaption}
\usepackage{amssymb}
\usepackage{amsmath}
\usepackage{MnSymbol}

\usetikzlibrary{external}
\usetikzlibrary{arrows,shapes,plotmarks,decorations.pathmorphing}
\usetikzlibrary{backgrounds,calc,positioning,fadings}

\usepgfplotslibrary{groupplots}

\tikzset{every picture/.style={node distance=2cm}}
\tikzset{>=stealth'} 
\tikzstyle{graphnode} = 
   [circle,draw=black!80!white,minimum size=20pt,text centered,text
     width=20,inner sep=0pt] 
\tikzstyle{var}   =[graphnode,fill=white]
\tikzstyle{obs}   =[graphnode,fill=black!80!white,text=white]
\tikzstyle{act}   =[rectangle,draw=black!80!white,text=white,minimum
size=22pt,text centered, text width=22pt,inner sep=0pt]
\tikzstyle{fac}   =[rectangle,draw=black!80!white,fill=TUgold,minimum size=5pt]
\tikzstyle{facprior} =[rectangle,draw=black!80!white,fill=black!80!white,text=white,minimum size=5pt]
\tikzstyle{edge}  =[draw=white,double=black!80!white,thick,-]
\tikzstyle{prior} =[rectangle, draw=black!80!white, fill=black!80!white, minimum size=
5pt, inner sep=0pt]
\tikzstyle{dirprior} = [circle, draw=black!80!white, fill=black!80!white, minimum
size=5pt, inner sep=0pt]

\newcommand{\JG}[1]{ #1}

\title{Probabilistic DAG Search}

\author[1]{\href{mailto:<julia.grosse@uni-tuebingen.de>?Subject=Probabilistic DAG Search}{Julia~Grosse}{}}
\author[2]{Cheng~Zhang}
\author[1,3]{Philipp~Hennig}
\affil[1]{%
    University of Tübingen\\
    Tübingen, Germany\\
}
\affil[2]{%
    Microsoft Research\\
    Cambridge, United Kingdom\\
}
\affil[3]{%
    Max-Planck-Institute for Intelligent Systems\\
    Tübingen, Germany\\
  }
  
\begin{document}
\maketitle

\begin{abstract}
Exciting contemporary machine learning problems have recently been phrased in the classic formalism of tree search---most famously, the game of Go. Interestingly, the state-space underlying these sequential decision-making problems often posses a more general latent structure than can be captured by a tree. In this work, we develop a probabilistic framework
to exploit a search space's latent structure and thereby share information across the search tree. The method is based on a combination of approximate inference in jointly Gaussian models for the explored part of the problem, and an abstraction for the unexplored part that imposes a reduction of complexity \emph{ad hoc}. 
We empirically find our algorithm to compare favorably to existing non-probabilistic alternatives in Tic-Tac-Toe and a feature selection application.

\end{abstract}
\section{Introduction}
Tree search is a natural formulation for problems that require planning and decision making. For historical reasons it is often studied in the domain of board games, 
but the scope of applications reaches wide beyond. Within machine learning, problems from reinforcement learning, active learning or experimental design can be phrased as tree search.  
More generally, tree search is for example applied to  combinatorial optimization \citep{sabharwal2012guiding} or scheduling problems \citep{pellier2010uct} in computer science.

The fundamental challenge for problems of this kind is the exponential complexity of the underlying search tree. In many cases the complexity can be reduced to some extent by collapsing nodes that represent identical problem states to a single node. In this work, we propose a method that additionally exploits similarity relations -- as a generalization of the identity relation -- between nodes to improve the search.

\begin{figure}
\centering
\begin{subfigure}{.25\textwidth}
  \centering
  \includegraphics[width=\linewidth]{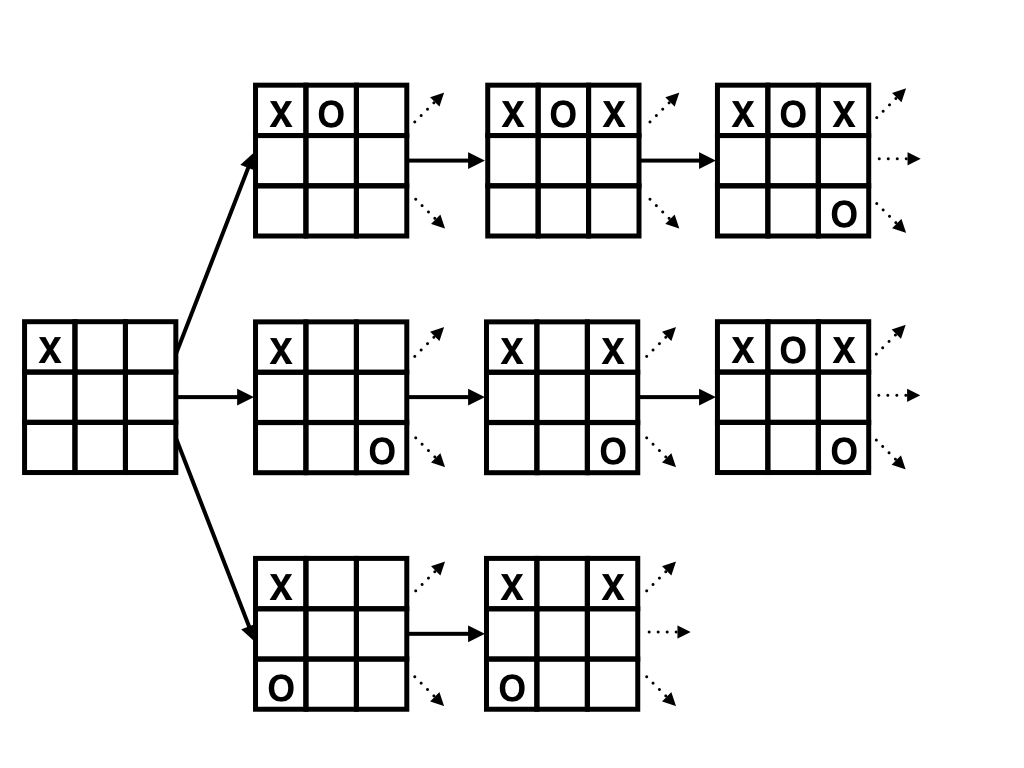}
  \caption{Search Tree for Tic-Tac-Toe.}
  \label{ttttree}
\end{subfigure}%
\begin{subfigure}{.25\textwidth}
  \centering
  \includegraphics[width=\linewidth]{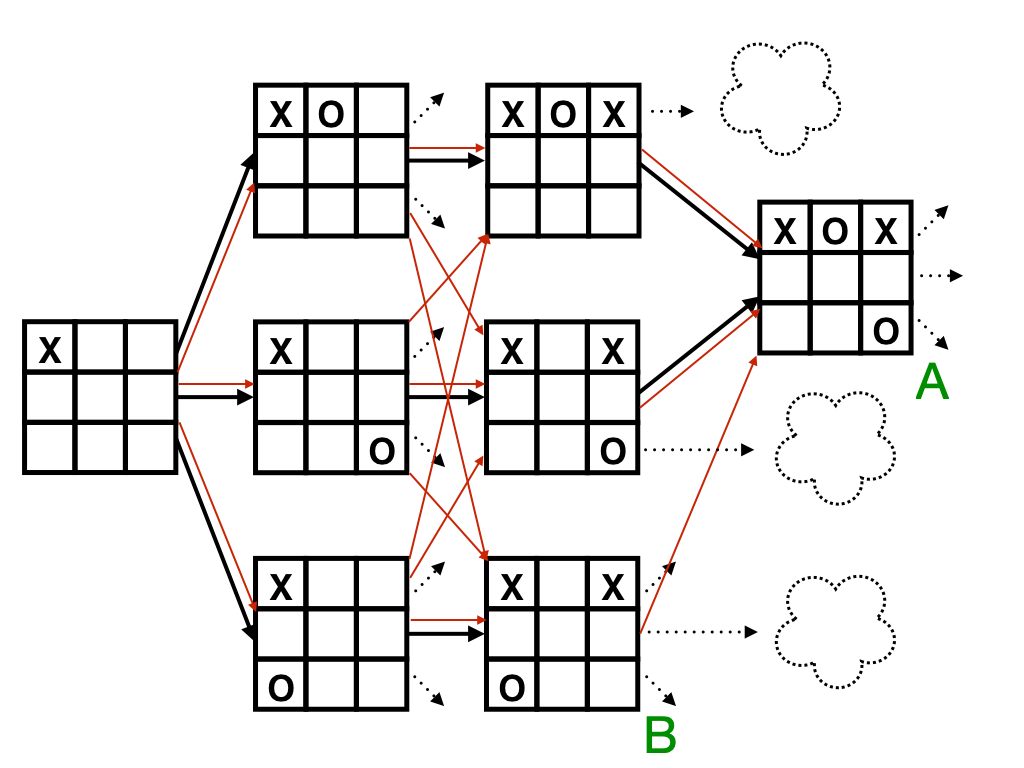}
  \caption{Search DAG for Tic-Tac-Toe.}
  \label{tttdag}
\end{subfigure}
\caption{The advantage of treating a search space as a DAG instead of a tree. (a) In a tree, identical states are represented by different nodes if reached by different move sequences. (b) In a DAG, such states are collapsed into one node. In the proposed model states with overlapping signs $(\times, \circ)$ will be related to each other even if they are not (yet) connected by a path, as indicated by the red lines. We assign a priori beliefs to the unexplored part of the DAG, illustrated by the clouds.}
\label{fig:ttttreedag}
\end{figure}

 As a didactic example, we will consider the search tree of the popular game Tic-Tac-Toe (Figure~\ref{fig:ttttreedag}). The nodes and paths represent positions and sequences of decisions made by the two players, respectively.
Different sequences of decisions sometimes lead to the same position---the paths overlap. While the tree has 58524 nodes, there are only 765 different board states in Tic-Tac-Toe. In the context of games this is known as \textit{move transposition}, but the same phenomenon also occurs in other decision making problems beyond the scope of games, e.g. in feature selection. Apart from representing the search space as a DAG, our method also includes relations between board states with overlapping signs, such as node A and B in Figure~\ref{fig:ttttreedag}(b). Observing a win by playing the game starting from state A also increases the estimated score of node B.\\

As a second, more realistic example (on account of its scale) we will study the problem of selecting a limited subset of pixels from MNIST images \cite{deng2012mnist}
to maximize classification accuracy of the images.
The greedy approach, considering the pixels in isolation and picking the most informative ones one after another, is evidently suboptimal: For example, neighbouring pixels could be selected that are informative in themselves, but together provide redundant information. 
However, testing all possible combinations is, well, \emph{combinatorially} hard. To find the best subset of $k=10$ pixels from the $N=28\times 28=784$ pixels of MNIST images, one has to search $\begin{pmatrix} N \\ k\end{pmatrix}\approx 2.28 \cdot 10^{22}$ possible combinations. One way to do so is to build a search tree starting with the empty subset as the root and iteratively adding the remaining features in each level. This tree has $\sum_{i=0}^{k}\frac{N!}{(N-i)!}\approx 8.29 \cdot 10^{28}$ states. For a feature selection problem, though, the order of features does not matter, so the search tree can be reduced to a directed acyclic graph (DAG), which has only $\sum_{i=0}^{k} \begin{pmatrix} N \\ i\end{pmatrix} \approx 2.31 \cdot 10^{22}$ nodes. This is still a large number, so further domain knowledge must be used to leverage \emph{``smoothness''}: 
For example, the values of nodes representing combinations of pixels, that have some pixels in common, can assumed to be similar to each other. 
The information collected to estimate the value of one of the nodes is thereby also useful to learn something about the values of the others.\\
We develop a probabilistic inference scheme for search problems on (smooth) DAGs, that allows information sharing across the search space, by extending an earlier approximate probabilistic inference framework for the optimal values in (adversarial game) trees \citep{hennig2010coherent}.

\subsection{Problem Definition and Notation}
We deal with the problem of finding an optimal sequence of decisions in a discrete space. The different sequences form the search space that is represented as a DAG (trees are a special case) $G = (X,E)$ with nodes $X$ and directed edges $E$. Each node $i \in X$ corresponds to a state $s_i$ of the search space. 
The \textit{root} of the DAG corresponds to the initial state. The \textit{children} of a node correspond to the states that can be reached from the \textit{parent's} state by making a decision. A path from the root to a \textit{leaf} (terminal node) describes a complete sequence of decisions. Leaf nodes are associated with a \textit{reward}, the value/utility of this sequence of decisions. 
The goal is to find an \textit{optimal path} through the DAG --- a path leading from the root to a leaf with maximal reward. 
We distinguish between two scenarios: in \textit{adversarial settings}, every second decision belongs to an opponent that tries to minimize the reward. Nodes at which the opponent takes the decision are therefore called MIN nodes. Levels in the tree with MIN nodes alternate with levels with MAX nodes. In \textit{non-adversarial settings}, there are only MAX nodes (or MIN nodes).
During search \emph{roll-outs} from a node can be requested. For a roll-out a path from this node to a leaf is descended, guided by a roll-out policy. We use a uniform random policy in this work.
The information available from a roll-out consists of the reached leaf $t$ and its reward $r_t$. 
Nodes have two kinds of values attached: a \textit{generative score g} and an \textit{optimal value v}. The generative score of a node models the obtained reward if a random path from this node to a leaf is followed. For the optimal value, it is instead assumed that an optimal decision is made at each node of such a path. 

\subsection{Background: Tree Search}

The most popular contemporary approach to tree search is the family of Monte Carlo Tree Search (MCTS) algorithms. MCTS involves building a growing \emph{search tree} by adding a new node in each iteration. Such an iteration begins with a descent from the root through the search tree, ending at a node that has not been visited before. From there begins a roll-out through the unexplored part of the tree. The reward observed at its end is backpropagated up to the root, improving a value estimate of the existing nodes, which in turn allows a more informative descent in the next step. A key implicit assumption underlying MCTS is that the search space is ``smooth'': The rewards at leaf nodes are not independent of each other, but those of nodes close to each other in the tree (i.e.~nodes that share a larger part of their ancestry) are more similar to each other \citep{pearl1984intelligent}.

Common forms of MCTS, such as the UCT (Upper Confidence Bound applied to trees) algorithm,
are based on PAC (probably approximately correct) point estimates \citep{auer1995gambling}. The main argument for PAC estimates is that they require little in the way of prior assumptions. This property is a double-edged sword, however, because it also \emph{prevents} the use of patently available prior information. The alternative is to construct an explicit probabilistic model for the tree and perform inference on it. The main problem with this latter approach is that it requires approximations, because inference on the maximum of correlated random variables is notoriously challenging.
An early work on such approximate methods for probabilistic tree search by \citet{hennig2010coherent} constructs a joint generative model for the values of nodes in a game tree both under random and under optimal play. Their construction begins with imposing a Gaussian generative model (based on a Brownian random walk) for the values of board positions \textit{under random play}. The motivation for this model is the intuition that each decision on the way from the root to a leaf changes the value of the position, the probability to win or lose, by a small amount, and this amount is iid.~(and assumed Gaussian for convenience) if both players act randomly. From this relatively simple Gaussian model, the op.cit.~then derive an inference scheme for the values under \textit{optimal play} as a recursive application of expectation propagation \citep{minka2001expectation,minka2001family}.\\

\section{Related Work} \label{relatedwork}
We integrate three aspects of tree search research: A probabilistic formulation of the search; search on DAGs; and generalization between related states.
In addition to the aforementioned probabilistic tree search by \cite{hennig2010coherent}, on which our methods are based, there are others described in \cite{tesauro2012bayesian}, \cite{baum1997bayesian} and \cite{stern2007learning}. In contrast to the method described above, the latter methods all model the optimal values of sibling nodes as independent of each other. 
There have been other (non-probabilistic) attempts to search on DAGs (Section \ref{searchspaceasdag}), and to generalize across states (Section \ref{generalizationbetweenrelatedstates}).

\subsection{Search Space as DAG} \label{searchspaceasdag}
The idea of treating the search space as a DAG instead of a tree proved valuable in chess (e.g. \cite{plaat1996exploiting}, \cite{warnock1988search}) and some other games (e.g. \cite{veness2007effective}, \cite{rasmussen2007template}). Because the predominant search algorithm of this era was alpha-beta search, much of the literature is concerned with this depth-first method. For contemporary MCTS, \cite{saffidine2012ucd}
suggest three ways how to incorporate data from roll-outs in the DAG setting, in which the ancestors traversed during the descent from the root to the currently explored node are a \emph{subset} of all the ancestors of the node. The simplest approach is to update only the ancestors in this subset with the usual UCT update rule. A second option is to update all ancestors. As a compromise between the first two options, one could also update the visited ancestors and non-visited ancestors who do not exceed a certain distance from the explored node. There are difficulties with all three options: In the first option, information is wasted. For the second option, \cite{saffidine2012ucd} provide a counterexample, where the search is likely not to converge, and in the third option, it is unknown how one should choose the distance. \cite{pelissier2019feature} describes another idea for MCTS on DAGs. Their work differs from ours in that they recursively update confidence intervals — in contrast to distributions — for the optimal values of interior nodes. Perhaps, the more interesting difference is how the confidence interval/distribution for the optimal value of a node at the boundary is constructed. The confidence interval is based solely on observations from this node, whereas in our work, the optimal values are based on the generative scores, which allows us to share information from observations made elsewhere.

\subsection{Generalization between Related States} \label{generalizationbetweenrelatedstates}
From the DAG perspective, nodes that represent identical states are collapsed into a single node. A generalization of this discrete form of similarity is the continuous-valued notion of a kernel / covariance. Above, we use this idea to construct a joint Gaussian model with a custom covariance function over the state space. This allows sharing observations over different parts of the DAG. 
A related, but less general idea is the so-called Rapid Action Value Estimation (RAVE), originally introduced for Go by \citet{gelly2007combining}, where RAVE assigns a global score to each of the $19\times 19= 361$ available board locations, which is updated wherever in the tree the move is selected. As illustrated above for the pixel selection scenario, the value of a decision often depends on the decisions made before. In order to take this aspect into account, one can also define \emph{local} RAVE-scores for each decision conditioned on others. Nodes are then selected based on a combination of the original value estimate and the RAVE scores. These scores have proven valuable in Go, but they are not straightforward to extend to other settings, where the set of available actions depends on the context of the state.
\definecolor{color0}{rgb}{0.945098039215686,0.435294117647059,0.125490196078431}
\definecolor{color1}{rgb}{0.843137254901961,0.6,0.133333333333333}
\definecolor{color2}{rgb}{0.250980392156863,0.337254901960784,0.631372549019608}
\definecolor{color3}{rgb}{1,0.498039215686275,0}
\definecolor{color4}{rgb}{0.94509803921,0.23529411764,0.12549019607}
\definecolor{color5}{rgb}{0.94509803921,0.23529411764,0.12549019607}

\begin{figure}[ht]
  \centering \scriptsize
   \tikzexternaldisable
  \begin{tikzpicture}
  	\node[var,draw=color3,text=color3] (vr) at (0.5,0.25) {$v_\emptyset$};\
  	\node[var,draw=color3,text=color3] (v1) at (-2.5,-0.75) {$v_{\{1\}}$} edge[->,draw=color3] (vr);
  	\node[var,draw=color3,text=color3] (v2) at (-0.5,-0.75) {$v_{\{x\}}$} edge[->,draw=color3] (vr);
  	\node[var,,draw=color3,text=color3] (v3) at (1.5,-0.75) {$v_{\{y\}}$} edge[->,draw=color3] (vr);
  	\node[var,draw=color3,text=color3] (v4) at (3.5,-0.75) {$v_{\{z\}}$} edge[->,draw=color3] (vr);
  	\node[color3] at (-1.5,-0.75) {$\cdots$};
  	\node[color3] at (0.5,-0.75) {$\cdots$};
  	\node[color3] at (2.5,-0.75) {$\cdots$};
  	\node[var,draw=color3,text=color3] (vd1) at (0.5,-1.75) {$v_{\{x,y\}}$} edge[->,color3] (v2) edge[->,color3] (v3);

  	\node[var] (r) at (0,0) {$g_\emptyset$};
  	\node[var] (c1) at (-3,-1) {$g_{\{1\}}$} edge[<-] (r) edge[->, out=120, in=120, looseness=2] (v1);
  	\node[var] (c2) at (-1,-1) {$g_{\{x\}}$} edge[<-] (r) edge[->, out=120, in=120, looseness=2] (v2) ;
  	\node[var] (c3) at (1,-1)  {$g_{\{y\}}$} edge[<-] (r) edge[->, out=120, in=120, looseness=2] (v3);
  	\node[var] (c4) at (3,-1)  {$g_{\{z\}}$} edge[<-] (r) edge[->, out=120, in=120, looseness=2] (v4);
  	\node at (-2,-1) {$\cdots$};
  	\node at (0,-1) {$\cdots$};
  	\node at (2,-1) {$\cdots$};

	\node[var] (d1) at (0,-2) {$g_{\{x,y\}}$} edge[<-] (c2) edge[<-] (c3) edge[->, out=120, in=120, looseness=2] (vd1);

	\node[obs] (o1) at (-3,-2.5) {$r_{\{1,...\}}$} edge[<-] (c1);
	\node[obs] (o2) at (0,-3.5) {$r_{\{x,y,z,.\}.}$} edge[<-] (d1) edge[->,color5, out=20, in=210, looseness=2] (c4);
	\node[obs] (o1) at (3,-2.5) {$r_{\{z,...\}}$} edge[<-] (c4);

	\node[var,minimum width=10pt,text width=10pt,draw=color2,text=color2] at (-2.33,-1.66) {$\Delta_1$} edge[->] (v1);
	\node[var,minimum width=10pt,text width=10pt,draw=color2,text=color2] at (0.66,-2.66) {$\Delta_{xy}$} edge[->] (vd1);
	\node[var,minimum width=10pt,text width=10pt,draw=color2,text=color2] at (3.66,-1.66) {$\Delta_z$} edge[->] (v4);

	\end{tikzpicture}

\caption{Generative model for feature selection. Each node in the DAG represents a feature bag. Three variables belong to each node: generative scores $g$ (Section \ref{generativemodel} and \ref{inferenceongenerativescores}) in black, optimal values $v$ (Section \ref{generativemodel} and \ref{inferenceonoptimalvalues}), and increments $\Delta$ (Section \ref{inference}). Roll-outs from a node result in the observation of a reward $r$ (Section \ref{generativemodel}), shown in black. The information from roll-outs is used to update the generative scores (Section \ref{inferenceongenerativescores}). It is shared across the search space, e.g. the observed reward in the roll-out from node $\{x,y\}$ influences the generative score of node $\{z\}$ (indicated by red line) if the terminal node also contains feature $z$. For nodes at the boundary, $v$ is inferred from $g$ and $v$ and for inner nodes, it is derived from the children’s $v$ (Section \ref{inferenceonoptimalvalues}).}
\label{fig:featuredag}
\end{figure}
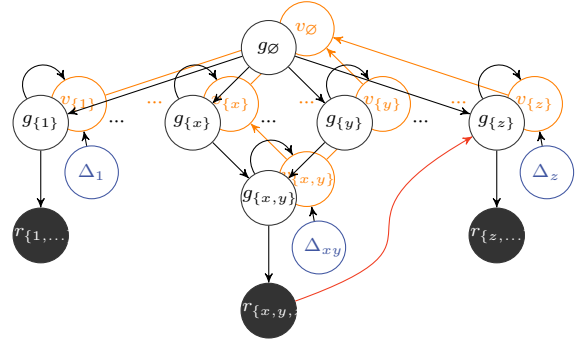

\section{Method}\label{method}
To construct a probabilistic method for search on DAGs, we extend the algorithmic structure of MCTS outlined above: The observed part of the DAG is traversed from the root to a node at the current boundary of the tracked DAG using a policy based on the beliefs over the optimal values. Once a node at the border is reached, this node is \textit{expanded}: The node's children are added to the DAG---if not already part of the DAG.
A roll-out from the node is then performed and the revealed information is used to update the beliefs over the generative and optimal values in the DAG. Doing so in a probabilistic fashion requires a generative model (Section \ref{generativemodel}), and an approximate algorithm for inference in it (Section \ref{inference}). Following the approach of \citet{hennig2010coherent}, we define the \emph{value}, the quantity to be optimized, in a recursive fashion by leveraging an easier-to-define Gaussian \emph{generative score} describing the evolution under random paths through the graph. Figure~\ref{fig:featuredag} provides an overview of the generative model in the context of a feature selection application. In this example, the generative score of a node, say node $\{x,y\}$, models the classification accuracy based on features $x$ and $y$ and $k-2$ additional, randomly selected features. For the node's optimal score $v_{x,y}$, the \emph{best} $k-2$ of the remaining features are selected instead. \\
A crucial part of the proposed probabilistic DAG search is that the information from a roll-out is not only used to update the estimates for the node and it's ancestors itself, but also for nodes in other parts of the search space, that are related to the leaf node reached by the roll-out. In Figure~\ref{fig:featuredag}, the roll-out from node \{x,y\} terminates in a leaf node, that also contains feature $z$. The reward $r_{x,y,z,..}$ collected there influences the generative score of node $\{z\}$. This is realised through a positive correlation of the modeled $g$-scores of nodes with shared features.

\subsection{Generative Model} \label{generativemodel}

\textbf{Definition.}
The joint prior distribution over the \emph{generative scores} $g_i$ over nodes $x_i \in X$ in the DAG $G =(X,E) $ is given by a multivariate Gaussian distribution
$p(g) = \mathcal{N}(g; \mu , c \cdot \Sigma)$ with arbitrary $\mu \in R^{|X|}$ and symmetric positive definite $\Sigma \in R^{|X| \times |X|}$.\\
The covariance $\Sigma_{ij}$ of the generative scores $g_i$ and $g_j$ should be defined such that it captures problem-specific relationship of the states represented by nodes $i$ and $j$. For example, in Tic-Tac-Toe, a plausible choice is the number of overlapping $\times$'s and $\circ$'s. In our second example, feature selection for MNIST images, we define the covariance as the shared number of selected pixels in the feature-bag represented by the two nodes ($+1$ on the diagonals to ensure positive semi-definiteness). \\
The scaling factor $c$ in the prior variance captures the range of possible rewards and depth of the tree. In our feature selection experiments we standardize the rewards at leaf nodes such that they have unit variance and zero mean. In this case, $c$ can be set to the inverse depth of the DAG and the prior mean $\mu$ to zero.

\textbf{Definition.}
The likelihood of an observation $r_t$ at a leaf node $t$ is defined as $p(r_t|g_t) = \mathcal{N}(r_t; g_t, \lambda)$ where $\lambda \in R $ is a small constant modeling noise in the observations. \\
For example, in game scenarios an observation $r_t$ consists of the outcome of the game encoded numerically and in the feature selection scenario an observation could be the classification accuracy. 

\textbf{Definition.} The \emph{optimal value} $v_t$ of a node is defined recursively: For a leaf node $t$, set $v_t=g_t$. For interior nodes in adversarial situations (where choices are made in alternating fashion between a MAX and a MIN player), the optimal value is the maximum or minimum of the children's optimal values:
\begin{align}
v_i = \begin{cases}
\max\{v_j | j \in \text{children}(i)\} & \text{if } i  \text{ is MAX}\\
\min\{v_j | j \in \text{children}(i)\} & \text{if } i \text{ is MIN}\\
\end{cases}
\end{align}
(for non-adversarial settings, every node is MAX).

\subsection{Inference} \label{inference}

The goal of our algorithm is to find a path through the DAG that leads to a leaf $t$ with globally maximal $v_t$. The end of the recursion in the definition of the optimal values are the leaf nodes. Since the DAG is only observed up to the current boundary and we only keep track of beliefs of observed nodes, we have to approximate the optimal values of the nodes at the boundary. This is where the generative scores come into play. We assume that the optimal value $v_i$ of a node $i$ at the boundary consists of its generative score $g_i$ and some increment $\Delta_i$:
\begin{align}
v_i = g_i + \Delta_i
\end{align}
The increment $\Delta_i$ of a node indicates how much we can improve the score of the node if the remaining steps are selected in an optimal manner in contrast to random selection. \cite{hennig2010coherent} show how to derive beliefs for these increments, prior to data acquisition. We take over their derivations, i.e. we treat the \textit{unexplored} part of the search space as a tree, and assume that the generative score $g_j$ of a child node $j$ can be obtained from the parent's generative score $g_i$ by adding a (scaled) Brownian step $\xi_j \sim \mathcal{N}(0,c)$ to the parent's generative score $g_i$:
\begin{align}
    p(g_{j}|g_{i}) = \mathcal{N}(g_{j}; g_{i}, c)
\end{align}
For leaf nodes $t$ the generative scores correspond to the optimal values, thus $\Delta_t=0$. Now, consider a node $i$ one level above the leaf nodes. If node $i$ is a MAX node and the best option is chosen, the score of the parent can be improved  by
\begin{align}
\Delta_i = \max_{j \in \text{children}(i)}\{\xi_j\}
\end{align}
For the $\Delta_i$ of nodes further up, one obtains recursively (for MAX nodes):
\begin{align}
\Delta_i = \max_{j \in \text{children}(i)}\{\Delta_j+ \xi_j\}
\end{align}
Assuming a constant branching factor, at least per level, the computation of the beliefs over the $\Delta$ can be simplified. With this assumption, the approximation for the $\Delta$ is the same for all nodes in the same level, i.e. it is sufficient to calculate the approximation for a single $\Delta$ per level and use copies for the sibling nodes for the subsequent step. The calculation effort for the $\Delta$'s is therefore not exponential, as it seems in their recursive definition, but linear in the depth of the tree/DAG.
The distribution over the maximum/minimum of Gaussian distributed random variables is not Gaussian anymore. However, there is a Gaussian approximation available \citep{Hennig2009}, that is summarized in Section \ref{approximationofthemaxmium}. We repeatedly apply this approximation to obtain a look-up table for the $\Delta$ prior to the search.\\
The Brownian steps from a parent node to its children, on which the definition of the $\Delta$ is based, are independent of each other in the tree model, i.e. the $\Delta$ of sibling nodes are not correlated. If the search space is actually a general DAG, the effective number of options is generally smaller than in a tree, because some paths reunite in later steps. In other words, the optimal increments $\Delta$ of sibling nodes are correlated. For example, think of the extreme case where all paths end in the same leaf node and the $\Delta$ would be perfectly correlated.

\subsubsection{Approximation of the Maximum of Gaussian Distributed Variables} \label{approximationofthemaxmium}
\begin{figure}\scriptsize
\centering
\tikzexternaldisable
\input{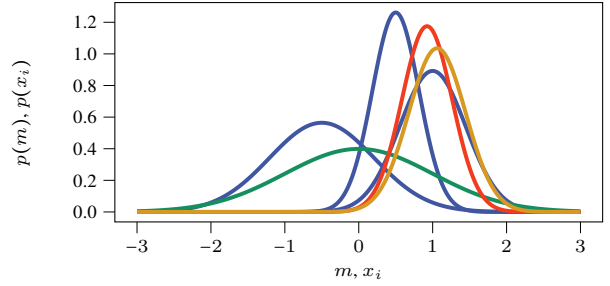}
\caption{Gaussian Approximation of the maximum of Gaussian distributed variables. The red and yellow lines show Gaussian approximations to the maximum $m$ of three independent Gaussian variables $x_1, x_2, x_3$ plotted in blue. The approximation plotted in red includes a Gaussian prior for the maximum (shown in green). The yellow one is obtained if no prior information on the maximum is included.} 
\label{fig:maxgauss}
\end{figure}

Consider the problem of determining the distribution of the maximum of a finite set of jointly Gaussian distributed variables, as required, inter alia, for the $\Delta$. Suppose, for the moment, one is interested in the maximum $m$ of only two variables $x_1$ and $x_2$. The knowledge $\mathcal{I}_{x}$ on $x_1$ and $x_2$ is expressed by a joint Gaussian distribution $p(x_1, x_2| \mathcal{I}_{x}) = \mathcal{N}(x; \mu, \Sigma)$. In some cases there might be prior information $\mathcal{I}_{0}$ on the maximum in form of a Gaussian belief $p(m |\mathcal{I}_{0}) = \mathcal{N}(m;\mu_{0}, \sigma^2_{0})$. Both sets of information are combined into a posterior-like distribution of $m$: \begin{multline}
p(m|\mathcal{I}_x, \mathcal{I}_0)\\
= Z^{-1} p(m|\mathcal{I}_0) \cdot \int \int p(x_1, x_2|m) p(x_1, x_2|\mathcal{I}_x) dx_1dx_2,
\end{multline}
where $Z$ is a normalization constant. 
The solution to this problem requires computing the probability for either of them to be the maximum, and the value of the maximum itself:
\begin{multline}
p(m|\mathcal{I}_x, \mathcal{I}_0)\\
= Z^{-1} \cdot \underbrace{p(m|\mathcal{I}_0) \int_{- \infty}^{\infty} \delta(x_1-m)\int_{-\infty}^{x_1} p(x_1, x_2|\mathcal{I}_x)dx_2dx_1}_{m \text{ is distributed like } x_1 \text{ if }x_1 > x_2}\\
+ Z^{-1} \cdot \underbrace{p(m|\mathcal{I}_0) \int_{- \infty}^{\infty} \delta(x_2-m)\int_{-\infty}^{x_2} p(x_1, x_2|\mathcal{I}_x)dx_1dx_2}_{m \text{ is distributed like } x_2 \text{ if }x_2 > x_1},
\end{multline}
where $\delta$ denotes the Dirac-delta function.
Solving the above integrals amounts to computing generalized error functions, which are intractable in the multivariate case \citep{Genz2009}. The resulting distribution is also not Gaussian. 
\cite{Hennig2009} derives the moments of $p(m|\mathcal{I}_x, \mathcal{I}_0)$ which can be used for moment matching this distribution with a Gaussian.
The resulting expressions are listed in appendix A.
The Gaussian approximation for the maximum of two variables can be easily extended to an iterative approximation scheme for the maximum of a finite set with more variables as follows. At first, the maximum over two of the variables is approximated. In each subsequent step, a further variable is added and the maximum of it and the previous maximum is approximated. An approximation for the minimum can be obtained in the same way since $\min_i\{x_i\} = - \max_i\{-x_i\}$. The computational complexity of approximating the extremal value of $b$ variables is
$O(b^2)$ if the variables are correlated and reduces to $O(b)$ if they can be assumed to be independent. Figure~\ref{fig:maxgauss} shows two approximations of the maximum of three independent Gaussian variables, one with the standard Gaussian as prior and another one without prior information (i.e. $\sigma_0 \rightarrow \infty$).

\subsubsection{Inference on Generative Scores}\label{inferenceongenerativescores}

During search, observations from roll-outs are used to iteratively update the distribution over the generative scores, which then also informs the optimal values. This is where the smoothness of the DAG is explicitly used. Since both prior and likelihood for $g$ are Gaussian, this task is straightforward: Given observations $O = \{({t_i}, r_{t_i}): i=1, ...,n \}$ from the first $n$ roll-outs, the posterior mean and covariance are obtained with the usual formulas for Gaussian regression:
\begin{align}
\mu'_i =  \mu_i + \Sigma_{iO}(\Sigma_{OO} + \lambda)^{-1} (r - \mu_{O}) 
\text{ and }\\
\Sigma'_{ij} = \Sigma_{ij} - \Sigma_{iO}(\Sigma_{OO} + \lambda)^{-1}\Sigma_{Oi} 
\end{align}
The matrix $\Sigma_{OO} \in \mathbb{R}^{n \times n}$ denotes the covariance between the $g$ scores of the observed leaf nodes $\{{t_i}: i=1,...,n\}$ and the vector $r \in \mathbb{R}^{n}$ contains the corresponding observed rewards $\{r_{t_i}: i=1,...,n\}$.
To simplify exposition we consider this generic joint Gaussian model in this paper and do not assume further simplifying structure to $\Sigma_{OO}$. This means that inference is of cubic cost in the size of the state-space in memory (not the full, potentially exponentially large problem!). We note in passing that a naive implementation would have cost $\mathcal{O}(N^4)$ because of repeated matrix inversions, but this can be reduced to $\mathcal{O}(N^3)$ through careful use of the matrix-inversion lemma. Of course the various means of speeding up Gaussian inference to quadratic and linear time complexity developed in the ML community over recent years can be applied more or less directly to our framework to alleviate this polynomial complexity issue \citep{snelson2006sparse,titsias2009variational,hensman2013gaussian}.

\subsubsection{Inference on Optimal Values}\label{inferenceonoptimalvalues}
 
While inference on the generative scores $g$ is easy, inference on the extremal values $v$ is non-analytic anywhere other than at the boundary of the graph: For boundary nodes, the optimal value is the sum of the generative score and $\Delta$ (Eq. 2). Since the beliefs over the generative scores and $\Delta$'s are Gaussian, their sum is also a Gaussian random variable:
\begin{align}
p(v_i) = \mathcal{N}(\mu_{g_i} + \mu_{\Delta_i}, \sigma^2_{g_i} + \sigma^2_{\Delta_i})
\end{align}
Yet, for the interior nodes, $v$ values are defined as the maximum or minimum of their children's $v$ values. 
We recursively use the approximation scheme from Section \ref{approximationofthemaxmium} to approximate the $v$ values of interior nodes based on the approximations of their children's $v$ values. 
The above scheme can actually account for correlations between $v$-values if they are known, but in our experiments we assumed independence between the optimal values. This is of course a strong, and formally incorrect simplification, but it is currently unclear how to correctly compute the correlation coefficients. 
Such assumption has the effect that maximas are overestimated and minimas are undererstimated, respectively. A simplistic, but in our experiments effective way to counteract this is to introduce a (Gaussian) prior for the maximum that acts as a regularizer. The selection of the hyperparameters of this prior can be motivated by the range of the rewards. Assuming standardized rewards, we use a standard normal distribution as prior for the optimal value. 

Inference on a node's optimal value requires beliefs over the optimal values of all children. This is why we suggest to add all children of a node to the DAG as soon as it is expanded. However, this effectively increases the size of the search DAG by the branching factor of the graph. For problems with a high branching size (such as the MNIST pixel-selection example below), this can be a computation and memory bottleneck. As a simple, lightweight approximation, we then calculate the optimal values by considering the optimal values of just the observed children and a single summary variable for the maximum of all the currently \emph{unobserved} children. We approximate the value of this unexplored option by the parent's generative score and the $\Delta$ one would get at the parents level assuming a branching factor equal to the number of the remaining unexplored children.

\subsubsection{Simplified Estimation of Optimal Values} \label{simplifiedestimationofoptimalvalues}
A further simplification of the inference on the optimal value $v_j$ at a parent's node can be achieved by replacing the EP inference scheme and instead approximating $v_j$ more ad-hoc, as a weighted sum of the children's values $v_j := \sum_{i=1}^K w_i v_i$ with higher weights $w_i$ for children with higher estimated $v_i$ value. Using the softmax function as a soft version of the maximum function we propose to set the weights $w_i$ to
\begin{align}
w_i = \frac{e^{- (\mu_{max}-\mu_i)/\sigma_{v_i}}}{\sum_{k=1}^K w_k}
\end{align}
with $ \mu_{max} = \max_{k} \mu_{v_k}$. The weight vector can be interpreted as a categorical distribution for each of the children nodes to be the best choice (the argmax). 
 Assuming independence between the children and constant weights, one obtains for the parent's mean $\mu_{v_j}$ and variance $\sigma^2_{v_j}$:
\begin{align}
\mu_{v_j} = \sum_{i} w_{i} \mu_{v_i} \text{ and } \sigma^2_{v_j} = \sum_{i}  w_i^2 \sigma^2_{v_i}
\end{align}
The experiments below suggest that this new update rule for the optimal values, although ad hoc, performs only slightly worse than full EP inference while making the implementation considerably easier.

\subsection{Policy} \label{policy}
The procedure outlined above yields local, approximately Gaussian distributions over of the optimal values $v$, which can be used to guide the search, as well as during evaluation. For exploratory \emph{search} we use a standard upper confidence bound \citep{auer1995gambling} policy $\pi$ to select nodes (for MAX nodes, and mutatis mutandis for MIN nodes):
\begin{align}
 \pi[j] = \arg \max_{i \in \text{children}(j)} \mu_{v_i} + \beta \sqrt{\log(n_j)\sigma_{v_i}^2 }  
\end{align}
The variable $n_j$ counts how often node $j$ was visited during the search
(cf.~Section \ref{experiments} on how the exploration parameter $\beta$ was set in the experiments). 

\subsection{Summary}
To wrap up, an iteration in probabilistic DAG search consists of the following steps: A new node is added to the explored DAG and roll-out is carried out. The reward from the roll-out is used to update the generative scores of all nodes on the boundary via a Gaussian regression. Marginals for the optimal values of boundary nodes are inferred by adding up the marginals for the generative scores $g$ and the pre-computed optimal increments $\Delta$. The optimal values of all of the remaining nodes in the explored part of the DAG are updated using the Gaussian approximation scheme. 
\section{Experiments} \label{experiments}
We test our method in three different settings:
The first experiment is on a synthetically generated DAG/tree, where the covariance structure of the ground truth is in full accordance with the generative model assumed by the probabilistic DAG search.
The second experiment, Tic-Tac-Toe, is an example for an adversarial search problem. 
The last one is a high-dimensional feature selection experiment on MNIST with a branching factor $b=784$, higher than (but not as deep as) that of the Go game tree. See Figure \ref{fig:featuredag} for how to formulate features selection as DAG search. We compare our algorithm both to the probabilistic tree search of \cite{hennig2010coherent}, as well as classic UCT
and UCT on DAGs (UCD) \citep{saffidine2012ucd}, as discussed in Section \ref{searchspaceasdag}. We use the simplest version of UCD, where only the traversed ancestors are updated.\\

\subsection{Synthetic search problem}
\begin{figure}[t!]
\centering \scriptsize
 \tikzexternaldisable
\input{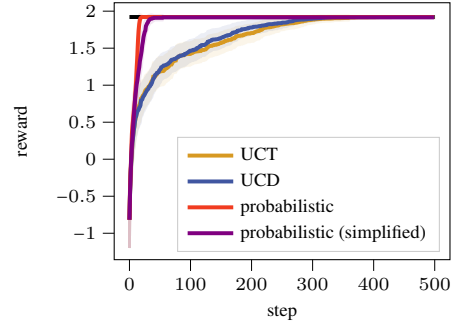}
  \caption{Comparison on synthetic DAG/tree. Average and standard deviation of reward from 100 repetitions. Black line indicates maximal reward.}
  \label{fig:synthetic}
 \end{figure}

In this first problem, the underlying search space is a DAG $G=(X,E)$ with nodes $X=\{x \subset F||x| \leq m\}$ and edges $E=\{(x_1, x_2)| x_2 = x_1 \cupdot {f} \in F\}$, where $F=\{1,...,N\}$. We impose a covariance $\Sigma \in \mathbb{R}^{|X| \times |X|}$ on the generative scores of nodes $x_i$ and $x_j$ defined as $\Sigma_{ij} = (|x_1 \cap x_2|+1)/(m+1)$. The ground-truth rewards at the leaf nodes are sampled from $\mathcal{N}(0,\Sigma)$. The ground truth covariance structure is known to the probabilistic versions. $N=15$ and $m=5$, i.e. the DAG has $|X|=4944$ nodes of which $3003$ are leaf nodes.
$\lambda$ was set to $10^{-4}$ for the probabilistic versions. $\beta \in \{0.01, 0.1, 1, \sqrt{2}, 10\}$ for all versions and Figure~\ref{fig:synthetic} shows the results from the best choice, respectively. It is clearly advantageous to use the covariance, whereas the reduction in number of nodes from tree to DAG has only a negligible effect on performance. We also note that, here, the simplified inference rule achieves nearly the same performance as the fully probabilistic version.

\subsection{Tic-Tac-Toe}

\tikzexternalenable
\begin{figure}\scriptsize
  \begin{center}
  \tikzexternaldisable
    	\input{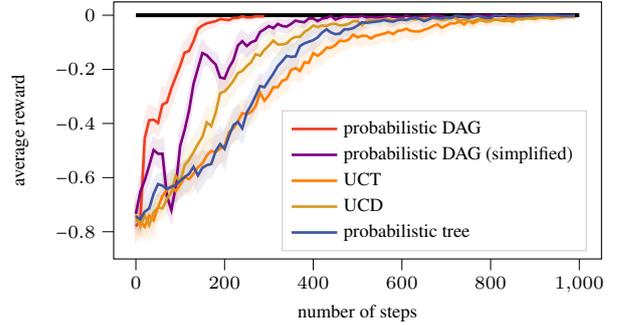}
  \end{center}
  \caption{Empirical performance on the Tic-Tac-Toe task. Average reward from 20 search runs. Standard deviations over 25 such repetitions (of 20 runs) as shaded regions.}
  \label{fig:ttt}
\end{figure}

The outcome of Tic-Tac-Toe is either a win ($r=1$), a draw ($r=0$) or a loss ($r=-1$). All search versions are evaluated after each step against an optimally playing MIN player. In this case the best outcome that the MAX player can reach is a draw. During evaluation the MAX player selects its move based on the maximum-a-posterior estimates of the $v$ values or selects its move randomly if reaches a previously unexplored node. 
We used the hyperparameters $c=0.5, \lambda=0.1$ for the probabilistic DAG and tree version and $\lambda=0.001$ for the softmax-based simplification (we did not perform extensive hyperparameter search, as these are relatively natural choices). The exploration constant is set to $\beta=1$ for all search versions.
Figure \ref{fig:ttt} shows the average reward for the MAX player with respect to the number of search steps. The probabilistic search on the DAG is able to use its similarity metric to outperform the other algorithms by a few hundred steps. Another factor for the better performance of the probabilistic DAG search could be that several nodes (all children) are added to the DAG in each search step. We also found that UCD worked better than UCT, supporting the expected gain from the reduced overall number of states. 
\JG{A comparison of the probabilistic DAG version using the fully probabilistic updates and the softmax-based simplification suggests that the lead of the probabilisic DAG version is not only due to the covariance structure, but also due to the probabilistic treatment of the optimal value estimation.}

 \begin{figure}\scriptsize
  \tikzexternaldisable
 \input{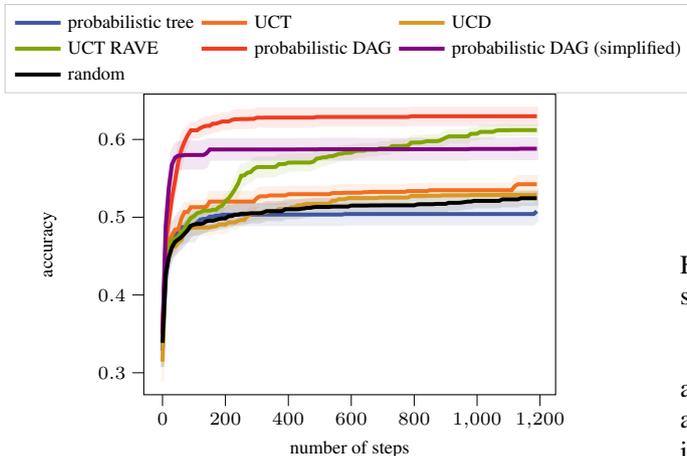}
 \caption{Empirical performance on the MNIST pixel selection task. Average reward from 10 rounds.}
 \label{fig:mnist-acc}
 \end{figure}

\subsection{Pixel Selection for MNIST Images}
Our final experimental task is to select a subset of 10 pixels from the MNIST images based on which the images are classified.
As a classifier for this task, we use a Convnet (cf.~Appendix B.1). The classifier is trained with images where all but 10 randomly selected pixels are masked. 1000 randomly selected images of the MNIST train set are held back as validation set. During the search, the classifier is evaluated on this set to obtain estimates of the accuracy. Before the search, 20 random roll-outs are performed and the standard deviation and mean of the observed accuracies are used to standardize the rewards. We use $\beta=0.5$ as exploration constant for all search versions and, since the classification accuracy can be evaluated to high precision, $\lambda=10^{-4}$ as noise parameter for the likelihood in the probabilistic methods. For the approximation of maxima and $\Delta$ we use a standard Gaussian prior. For this high-dimensional feature selection problem, UCT explores the search space almost uniformly, due to the high branching factor. \cite{gaudel2010feature} propose a heuristic based on RAVE scores in order to make UCT applicable for large scale feature selection problems. A detailed description of this heuristic can be found in Appendix B.2. 
Figure~\ref{fig:mnist-acc} shows the highest validation accuracy obtained by previous evaluations of leaf nodes. We observe a difference in performance between the search versions that generalize across the state -- probabilisitic DAG, probabilistic DAG (simplified), UCT-RAVE -- and the remaining ones. This indicates that the exploitation of similarity relations among states was more important than just a reduction in the number of states based on identity relations and even essential to learn anything on this problem. We also find a headstart of the probabilistic DAG version(s) as compared to UCT-RAVE and an advantage of the full-probabilistic update over the simplified, softmax-based updates.
\begin{figure}
\centering
\begin{minipage}{.25\textwidth}
\centering
 \tikzexternaldisable
\begin{tikzpicture}

\definecolor{color0}{rgb}{0.945098039215686,0.435294117647059,0.125490196078431}
\definecolor{color1}{rgb}{0.843137254901961,0.6,0.133333333333333}
\definecolor{color2}{rgb}{0.250980392156863,0.337254901960784,0.631372549019608}
\definecolor{color3}{rgb}{1,0.498039215686275,0}
\definecolor{color4}{rgb}{0.94509803921,0.23529411764,0.12549019607}

\begin{groupplot}[group style={group size=3 by 3}]
\nextgroupplot[
height=4.5cm,
minor xtick={},
minor ytick={},
tick pos=both,
width=4.5cm,
xmin=-0.5, xmax=27.5,
xtick=\empty,
y dir=reverse,
ymin=-0.5, ymax=27.5,
ytick=\empty
]
\addplot graphics [includegraphics cmd=\pgfimage,xmin=-0.5, xmax=27.5, ymin=27.5, ymax=-0.5] {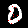};
\node[anchor=west] (source) at (axis cs:8,2){};
\node (destination) at (axis cs:13,7){};
\draw[->, draw=yellow](source)--(destination);
\draw[draw=yellow](source)--(destination);
\draw (axis cs:7,3) node[
  scale=0.9,
  anchor=base west,
  text=yellow,
  rotate=0.0
]{0};
\node[anchor=west] (source) at (axis cs:20,5){};
\node (destination) at (axis cs:14,7){};
\draw[->, draw=yellow](source)--(destination);
\draw[draw=yellow](source)--(destination);
\draw (axis cs:20,5) node[
  scale=0.9,
  anchor=base west,
  text=yellow,
  rotate=0.0
]{1};
\node[anchor=west] (source) at (axis cs:4,11){};
\node (destination) at (axis cs:10,11){};
\draw[->, draw=yellow](source)--(destination);
\draw[draw=yellow](source)--(destination);
\draw (axis cs:3,11) node[
  scale=0.9,
  anchor=base west,
  text=yellow,
  rotate=0.0
]{2};
\draw (axis cs:13,14) node[
  scale=0.9,
  anchor=base west,
  text=yellow,
  rotate=0.0
]{3};
\node[anchor=west] (source) at (axis cs:23,15){};
\node (destination) at (axis cs:17,15){};
\draw[->, draw=yellow](source)--(destination);
\draw[draw=yellow](source)--(destination);
\draw (axis cs:23,15) node[
  scale=0.9,
  anchor=base west,
  text=yellow,
  rotate=0.0
]{4};
\draw (axis cs:13,16) node[
  scale=0.9,
  anchor=base west,
  text=yellow,
  rotate=0.0
]{5};
\node[anchor=west] (source) at (axis cs:3,16){};
\node (destination) at (axis cs:10,19){};
\draw[->, draw=yellow](source)--(destination);
\draw[draw=yellow](source)--(destination);
\draw (axis cs:3,16) node[
  scale=0.9,
  anchor=base west,
  text=yellow,
  rotate=0.0
]{6};
\node[anchor=west] (source) at (axis cs:2,20){};
\node (destination) at (axis cs:9,20){};
\draw[->, draw=yellow](source)--(destination);
\draw[draw=yellow](source)--(destination);
\draw (axis cs:2,20) node[
  scale=0.9,
  anchor=base west,
  text=yellow,
  rotate=0.0
]{7};
\node[anchor=west] (source) at (axis cs:4,25){};
\node (destination) at (axis cs:11,23){};
\draw[->, draw=yellow](source)--(destination);
\draw[draw=yellow](source)--(destination);
\draw (axis cs:4,25) node[
  scale=0.9,
  anchor=base west,
  text=yellow,
  rotate=0.0
]{8};
\node[anchor=west] (source) at (axis cs:22,20){};
\node (destination) at (axis cs:18,20){};
\draw[->, draw=yellow](source)--(destination);
\draw[draw=yellow](source)--(destination);
\draw (axis cs:22,20) node[
  scale=0.9,
  anchor=base west,
  text=yellow,
  rotate=0.0
]{9};
\end{groupplot}

\end{tikzpicture}
  \label{fig:pixels}
\end{minipage}%
\begin{minipage}{.25\textwidth}
 \tikzexternaldisable
\begin{tikzpicture}

\begin{groupplot}[group style={group size=3 by 3,horizontal sep=2pt,vertical sep=2pt}, width=10, height=10]
\nextgroupplot[
height=2.5cm,
minor xtick={},
minor ytick={},
tick pos=both,
width=2.5cm,
xmin=-0.5, xmax=27.5,
xtick=\empty,
y dir=reverse,
ymin=-0.5, ymax=27.5,
ytick=\empty
]
\addplot graphics [includegraphics cmd=\pgfimage,xmin=-0.5, xmax=27.5, ymin=27.5, ymax=-0.5] {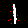};

\nextgroupplot[
height=2.5cm,
minor xtick={},
minor ytick={},
tick pos=both,
width=2.5cm,
xmin=-0.5, xmax=27.5,
xtick=\empty,
y dir=reverse,
ymin=-0.5, ymax=27.5,
ytick=\empty
]
\addplot graphics [includegraphics cmd=\pgfimage,xmin=-0.5, xmax=27.5, ymin=27.5, ymax=-0.5] {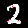};

\nextgroupplot[
height=2.5cm,
minor xtick={},
minor ytick={},
tick pos=both,
width=2.5cm,
xmin=-0.5, xmax=27.5,
xtick=\empty,
y dir=reverse,
ymin=-0.5, ymax=27.5,
ytick=\empty
]
\addplot graphics [includegraphics cmd=\pgfimage,xmin=-0.5, xmax=27.5, ymin=27.5, ymax=-0.5] {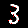};

\nextgroupplot[
height=2.5cm,
minor xtick={},
minor ytick={},
tick pos=both,
width=2.5cm,
xmin=-0.5, xmax=27.5,
xtick=\empty,
y dir=reverse,
ymin=-0.5, ymax=27.5,
ytick=\empty
]
\addplot graphics [includegraphics cmd=\pgfimage,xmin=-0.5, xmax=27.5, ymin=27.5, ymax=-0.5] {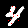};

\nextgroupplot[
height=2.5cm,
minor xtick={},
minor ytick={},
tick pos=both,
width=2.5cm,
xmin=-0.5, xmax=27.5,
xtick=\empty,
y dir=reverse,
ymin=-0.5, ymax=27.5,
ytick=\empty
]
\addplot graphics [includegraphics cmd=\pgfimage,xmin=-0.5, xmax=27.5, ymin=27.5, ymax=-0.5] {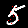};

\nextgroupplot[
height=2.5cm,
minor xtick={},
minor ytick={},
tick pos=both,
width=2.5cm,
xmin=-0.5, xmax=27.5,
xtick=\empty,
y dir=reverse,
ymin=-0.5, ymax=27.5,
ytick=\empty
]
\addplot graphics [includegraphics cmd=\pgfimage,xmin=-0.5, xmax=27.5, ymin=27.5, ymax=-0.5] {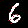};

\nextgroupplot[
height=2.5cm,
minor xtick={},
minor ytick={},
tick pos=both,
width=2.5cm,
xmin=-0.5, xmax=27.5,
xtick=\empty,
y dir=reverse,
ymin=-0.5, ymax=27.5,
ytick=\empty
]
\addplot graphics [includegraphics cmd=\pgfimage,xmin=-0.5, xmax=27.5, ymin=27.5, ymax=-0.5] {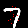};

\nextgroupplot[
height=2.5cm,
minor xtick={},
minor ytick={},
tick pos=both,
width=2.5cm,
xmin=-0.5, xmax=27.5,
xtick=\empty,
y dir=reverse,
ymin=-0.5, ymax=27.5,
ytick=\empty
]
\addplot graphics [includegraphics cmd=\pgfimage,xmin=-0.5, xmax=27.5, ymin=27.5, ymax=-0.5] {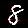};

\nextgroupplot[
height=2.5cm,
minor xtick={},
minor ytick={},
tick pos=both,
width=2.5cm,
xmin=-0.5, xmax=27.5,
xtick=\empty,
y dir=reverse,
ymin=-0.5, ymax=27.5,
ytick=\empty
]
\addplot graphics [includegraphics cmd=\pgfimage,xmin=-0.5, xmax=27.5, ymin=27.5, ymax=-0.5] {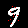};
\end{groupplot}

\end{tikzpicture}
\label{fig:classes}
\end{minipage}
\caption{Final set of selected pixels, show on randomly selected images from the MNIST test set.}  
\label{fig:MNIST-pixels}
\end{figure}
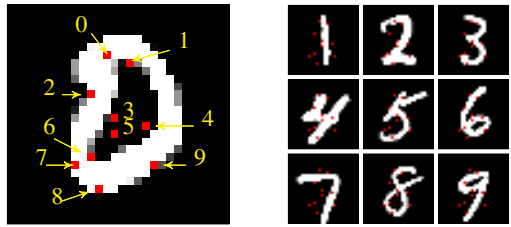

\begin{figure}[t!]
\centering \scriptsize
 \tikzexternaldisable
\input{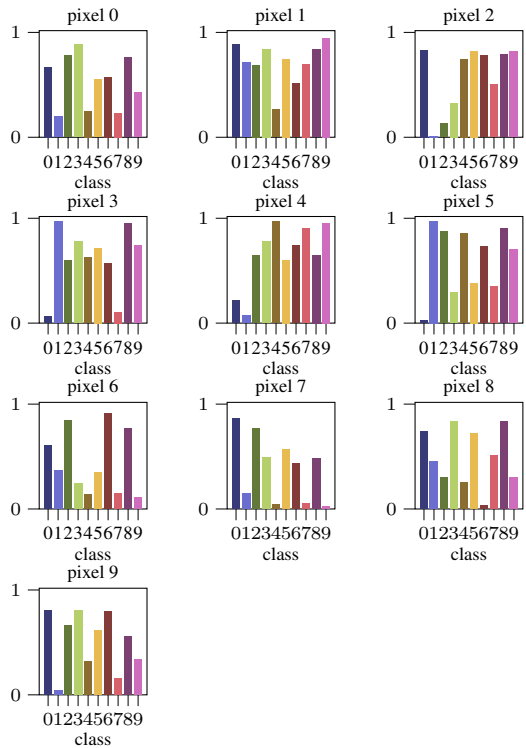}
  \caption{Proportion of images per class with positive value for the 10 selected pixels in Figure~\ref{fig:MNIST-pixels} (labelled accordingly).}
  \label{fig:mnist-bars}
 \end{figure}

Figure~\ref{fig:MNIST-pixels} shows 10 pixels selected by our method after 1200 steps. Figure~\ref{fig:mnist-bars} shows, for each class and each pixel, the proportion of images in the test dataset for which the respective pixel has a positive value. They give an intuition for their discriminative power. For example, pixel 3 separates classes 0 and 7 well from classes 1 and 8.

\section{Conclusion}

We have presented a probabilistic search algorithm for problems whose state-spaces can be represented by DAGs. A benefit of this formulation over tree search, beyond the obvious reduction in the size of the state-space itself, is that it allows for an exploitation of similarities between states. The key ingredients of our method are an approximate inference scheme for extremal values, which we constructed as an instance of expectation propagation, and an ad-hoc prior for the optimal value of unexplored parts of the search space, for which we extended earlier work on tree-structured problems.
Our experiments in the adversarial setting of Tic-Tac-Toe and the non-adversarial case of feature selection demonstrate not just good performance but also scalability. \JG{In addition, we examined a simplified version for the inference of optimal values as opposed to the one in \cite{hennig2010coherent}, that is based on the softmax function as a relaxation of the argmax function.}
 While it is relatively straightforward to include correlations between child nodes in the approximate inference scheme, it is challenging to construct a model to identify such correlations across the state space, a task that we reserve for future work.

\begin{contributions} 
    All authors contributed to the idea. Julia Grosse implemented the experiments and wrote the paper. 
    Philipp Hennig and Cheng Zhang made smaller edits and Philipp Hennig created Figure \ref{fig:featuredag}.
\end{contributions}

\begin{acknowledgements} 
    This work was supported by Microsoft Research through its PhD Scholarship Programme.
    The authors thank the International Max Planck Research School for Intelligent Systems (IMPRS-IS) for supporting Julia Grosse by non-financial means. The authors gratefully acknowledge financial support by the European Research through ERC StG Action 757275 / PANAMA; the DFG Cluster of Excellence "Machine Learning - New Perspectives for Science", EXC 2064/1, project number 390727645; the German Federal Ministry of Education and Research (BMBF) through the Tübingen AI Center (FKZ: 01IS18039A); and funds from the Ministry of Science, Research and Arts of the State of Baden-Württemberg.
\end{acknowledgements}

\bibliography{grosse_540}

\end{document}


\onecolumn
\maketitle

\section{A Gaussian Approximation of the Maximum of Gaussian Distributed Variables}
Here we give the moments used for the approximations in Section \ref{method} of the paper for the sake of completeness. Their derivations and an extended analysis of the approximation quality of the moment matching can be found in \cite{Hennig2009}.

Assume two jointly Gaussian distributed variables $x_1$ and $x_2$ with mean $\mu = \begin{pmatrix} \mu_1\\ \mu_2\end{pmatrix}$ and covariance $\Sigma = \begin{pmatrix}
\sigma_1^2 & \varrho \sigma_1 \sigma_2\\
\varrho \sigma_1 \sigma_2 & \sigma_2^2
\end{pmatrix}$, where $\varrho$ is their correlation. Given prior information $\mathcal{N}(\mu_0, \sigma_0^2)$ on their maximum $m$, the moments $\mu_{m}, \sigma_{m}^2$of the distribution of $m$ are:
\begin{align}
    \mu_{m} &= w_1 \biggl[ \mu_{c1} + \sigma_{c1} \frac{b_1}{a_1} \frac{\phi(k_1)}{\Phi(k_1)}\biggr] + w_2 \biggl[\mu_{c2} + \sigma_{c2} \frac{b_2}{a_2} \frac{\phi(k_2)}{\Phi(k_2)}\biggr]\\
    \begin{split}
    \sigma_m^2 &= w_1 \biggl \{ [2\mu_{c1}\sigma_{c1}] + \biggl[ 2\mu_{c1}\sigma_{c1} \frac{b_1}{a_1} - k_1 \sigma_{c1}^2 \frac{b_1^2}{a_1^2}\biggr] \frac{\phi(k_1)}{\Phi(k_1)} \biggr \} \\
    &+
    w_2 \biggl \{ [\mu_{c2}^2 + \sigma_{c2}^2] + \biggl[2\mu_{c2}\sigma_{c2} \frac{b_2}{a_2} - k_2 \sigma_{c2}^2 \frac{b_2^2}{a_2^2} \frac{\phi(k_2)}{\Phi(k_2)} \biggr] \frac{\phi(k_2)}{\Phi(k_2)}\biggr \} \\
    &- \mu_m^2
    \end{split},
\end{align}
where $\phi$ is the standard normal probability density function and $\Phi$ is the standard normal cumulative distribution function. The remaining terms are defined by:
\begin{align}
w_1 &= Z^{-1} \mathcal{N}(\mu_{0}; \mu_{1}, \sigma_{0}^2 + \sigma_1^2) \Phi(k_1) \\
w_2 &= Z^{-1} \mathcal{N}(\mu_{0}; \mu_{2}, \sigma_{0}^2 + \sigma_2^2) \Phi(k_2)\\
a_1 &= [\sigma_{1}^2 \sigma_{2}^2 (1-\varrho^2) + (\sigma_{1}-\varrho \sigma_{2})^2 \sigma_{c1}^2]^{1/2} \\
a_2 &= [\sigma_{1}^2 \sigma_{2}^2 (1-\varrho^2) + (\sigma_{2}-\varrho \sigma{1})^2 \sigma^2_{c2}]^{1/2}\\
b_1 &= \sigma_{c1}(\sigma_{1}-\varrho \sigma_{2}) \\
b_2 &= \sigma_{c2}(\sigma_{2} - \varrho \sigma_{1})\\
\sigma_{c1}^2 &= \frac{\sigma_{1}^2 \sigma_{0}^2}{\sigma_{c1}^2 + \sigma_{0}^2} \\
\mu_{c1} &= \biggl(\frac{\mu_{1}}{\sigma^2_{1}} + \frac{\mu_0}{\sigma_{0}^2} \biggr) \sigma_{c1}^2\\
Z &= \mathcal{N}(\mu_{0}; \mu_{1}, \sigma_{0}^2 + \sigma_{1}^2) \Phi(k_1) + \mathcal{N}(\mu_0; \mu_{2}, \sigma_{0}^2 + \sigma_{2}^2) \Phi(k_2)\\
k_1 &= \frac{(\sigma_{1} - \varrho \sigma_{2}) \mu_{c1}-\sigma_{1}\mu_{c1}-\sigma_{1}\mu_{2} + \varrho \sigma_{2}\mu_{1}}{[\sigma_{1}^2\sigma_{2}^2(1-\varrho^2) + (\sigma_{1}-\varrho \sigma_{2})^2\sigma_{c1}^2]^{1/2}}\\
k_2 &= \frac{(\sigma_{2} - \varrho \sigma_{1})\mu_{c2} - \sigma_{2}\mu_{1} + \varrho \sigma_{1}\mu_{2}}{[\sigma_{1}^2 \sigma_{2}^2(1-\varrho^2) + (\sigma_{2}-\varrho \sigma_{1})^2 \sigma_{c2}^2]^{1/2}}
\end{align}

\section{Additional Experimental Details}

This section provides additional experimental details for the MNIST pixel selection experiment. The tree/dag search methods can be combined with arbitrary classifiers to evaluate the reward at terminal nodes as long as it can handle partial input. We use a simple, fast Convnet (see Section B.1). The heuristic used for the UCT in MNIST experiment is given in Section B.2.

\subsection{Convnet for MNIST Experiment}
 The Convnet is built from two convolutional layers of size $32\times3\times3$ and $64\times3\times3$ with stride 1 and zero padding. The second convolutional layer is followed by $2\times2$ max pooling and two fully connected layers of size $9216\times128$ and $128\times10$. The inner layers use ReLUs and the softmax function is applied to the output layer. 
The first convolutional layer is followed by a 0.25 dropout layer and the first fully connected layer by a 0.5 dropout layer. Images are normalized during training and evaluation such that the pixel values are zero mean and unit standard deviation. During training all but randomly selected pixels are turned off, i.e. set to zero. We use stochastic gradient descent with the adaptive learning rate method (ADADELTA) and parameters  $\epsilon = 10^{-6}$, $\rho = 0.9$. The learning rate is set to $\lambda=1$in the beginning and is reduced by a factor of $\gamma=0.7$ after each epoch. The net is trained for 30 epochs with a batch size of 64. 

\subsection{RAVE heuristic for UCT}
UCT in its original form selects new nodes based on the UCB formula. At a MAX node $j$, the child $i$ that maximizes
\begin{align}
\mu_j + \beta \sqrt{\frac{ln(n_i)}{n_j}}
\end{align}
is chosen, where $\mu_j$ is the empirical mean of rewards from roll-outs passing through node $j$. The variables $n_j$ and $n_i$ count the visits of node $i$ and $j$ and are used together with hyperparameter $\beta$  to control the exploration. 
The RAVE-heuristic, orginally proposed by \cite{gelly2007combining} and and adapted to feature selection settings by \cite{gaudel2010feature}, combines the above selection policy with  local and global RAVE scores. 
The global RAVE score of pixel $p$ $g-\text{RAVE}_p$ indicates the global relevance of $p$ and is calculated as the average of the observed rewards at terminal nodes containing $p$:
\begin{align}
g-\text{RAVE}_p = \frac{1}{|T_p|} \sum_{i \in T_{p}} r_i
\end{align}
where $T_{p}$ denotes the set of observed terminal nodes that contain pixel $p$.
The local RAVE-score represents the importance of pixel $p$ at a node $i$, conditioned on the other pixels from node$i$. Here, the average is taken only from roll-outs that passed through node $i$. 
\begin{align}
l-\text{RAVE}_{i, p} = \frac{1}{|\{j \in T_p| i \rightsquigarrow j \}|} \sum_{j \in T_{p}, i \rightsquigarrow j } r_j
\end{align}
The modified score trades-off the original UCB term, the global RAVE score and the local one. It further includes a restriction of the exploration term in order to handle problems with high branching factor. 
The final score of a child $j$ reached from parent $i$ by adding pixel $p$ is defined as: 
\begin{align}
\begin{split}
(1- \alpha) \cdot \mu_{j} + \alpha \cdot ((1-\beta) \cdot l-\text{RAVE}_{i, p} + \beta \cdot g-\text{RAVE}_p) + \sqrt{\frac{c_{1} \ln(n_i)}{n_j} \min \biggl( \frac{1}{4}, \sigma_{j}^2 + \sqrt{\frac{2 \ln(n_i)}{n_j}}\biggr) }
\end{split}
\end{align}
$\sigma_j^2$ denotes the empirical variance of rewards observed from roll-outs passing through node $j$.
The trade-off is controlled by parameters $\alpha$ and $\beta$:
\begin{align}
\alpha = \frac{c_2}{c_2 + n_{j}}\\
\beta = \frac{c_3}{c_3 + n_{l}}
\end{align}
 $n_{l}$ denotes the number of observations that are taken into account for the calculation of the $l$-RAVE score. In the beginning of the search, the RAVE scores that are biased but calculated from more trials count more. In the course of time, their impact decreases and  the mean estimate $\mu_{j}$ is valued more as it is unbiased, but requires more trials to be reliable. 
The heuristic involves three hyperparameters $c_1, c_2, c_3$ that were set to $c_1= 10^{-4}, c_2= 10^{4}$ and $c_3= 10^{4}$, respectively.